 \documentclass[runningheads,a4paper]{llncs} \usepackage[T1]{fontenc}

\usepackage{amsmath} 
\usepackage{amsfonts} 
\usepackage{amssymb} 
\usepackage{graphicx}

\usepackage{hyperref}

\usepackage{color}
\usepackage{multirow}

\usepackage{chngcntr} 
\counterwithout{figure}{section} 
\counterwithout{figure}{subsection}

 \urldef{\mailsa}\path|{michal.nowicki, jan.wietrzykowski}@put.poznan.pl| \newcommand{\keywords}[1]{\par\addvspace\baselineskip \noindent\keywordname\enspace\ignorespaces#1}

  \usepackage{everypage} 
\newcommand\atxy[3]{%
 \AddThispageHook{\smash{\hspace*{\dimexpr+#1\relax}%
  \raisebox{\dimexpr+\voffset-#2\relax}{#3}}}}

\begin{document}

\atxy{1cm}{-1cm}{\parbox{\textwidth}{\footnotesize{The final publication is available at Springer via \url{http://dx.doi.org/10.1007/978-3-319-54042-9\_57}}}}

\mainmatter  

\title{Low-effort place recognition with WiFi fingerprints using deep learning}

\titlerunning{Low-effort place recognition with WiFi fingerprints using deep learning}

\author{Micha\l{} Nowicki\thanks{Corresponding author} \and Jan Wietrzykowski} %
\authorrunning{Micha\l{} Nowicki\thanks{Corresponding author} \and Jan Wietrzykowski} 

\institute{Institute of Control and Information Engineering,\\            Pozna\'n University of Technology\\            ul. Piotrowo 3A, 60-965 Pozna\'n, Poland\\ \mailsa\\}


\toctitle{Lecture Notes in Computer Science} 
\tocauthor{Authors' Instructions} 
\maketitle

 \begin{abstract}

Using WiFi signals for indoor localization is the main localization modality of the existing personal indoor localization systems operating on mobile devices. WiFi fingerprinting is also used for mobile robots, as WiFi signals are usually available indoors and can provide rough initial position estimate or can be used together with other positioning systems. Currently, the best solutions rely on filtering, manual data analysis, and time-consuming parameter tuning to achieve reliable and accurate localization. In this work, we propose to use deep neural networks to significantly lower the work-force burden of the localization system design, while still achieving satisfactory results. Assuming the state-of-the-art hierarchical approach, we employ the DNN system for building/floor classification. We show that stacked autoencoders allow to efficiently reduce the feature space in order to achieve robust and precise classification. The proposed architecture is verified on the publicly available UJIIndoorLoc dataset and the results are compared with other solutions.

 \end{abstract}

 \section{Introduction}

Indoor localization is a challenging task and there exist no universal solution for all possible applications. External infrastructure, such as networked cameras, is efficient for localization in limited areas~\cite{eaai01}. In large buildings the most precise agent's pose estimates are obtained with laser scanners~\cite{psamcs}, passive cameras~\cite{orbslam} or active RGB-D sensors~\cite{ecmr13}. Using these sensors it is possible to simultaneously localize the agent and build a map of the environment, solving the Simultaneous Localization and Mapping (SLAM) problem. Unfortunately, laser scanners are expensive, whereas processing camera or RGB-D images is computationally demanding, and requires a sophisticated processing pipeline to achieve satisfactory results \cite{iciar}. In SLAM we assume that no {\em a priori} information about the building structure is available. However, in practical applications building floor plans can be often gathered prior to the operation of the localization system. Having an access to {\em a priori} map enables us to use WiFi signal information for indoor localization. Nowadays, WiFi networks are ubiquitous in public buildings, offices, shopping malls, etc. Moreover, almost every mobile robot is equipped with a WiFi adapter used for connecting with the Internet or for remote operation. These adapters are also commonly available in mobile phones and tablets that may be used for personal indoor localization \cite{mobicase}. Therefore, WiFi information can be exploited to provide rough, global position estimates, without additional costs of exteroceptive sensors. One important issue that prevents wider adoption of this solution for indoor localization is the need to survey the entire environment for a WiFi signal strength map prior to the operation of the localization system. Thus, in this paper, we investigate if deep learning -- a recent and powerful machine learning paradigm can provide a global location recognition solution from WiFi data on a sparse map of scans, and at a significantly reduced effort for manual tuning.

\section{Related work}

Indoor localization with WiFi for mobile robots is an old idea~\cite{wifirobot1}, that was proved to be effective in systems using particle filtering approaches and joining WiFi localization with odometry readings~\cite{wifirobot2010}.  Possible application environments include buildings with existing WiFi infrastructure where other solutions are too expensive or yet not reliable enough, i.e. for refinery inspection~\cite{wifirobot2015}.  Although the most advanced WiFi systems are used for personal indoor localization with mobile devices (smartphones) in order to provide efficient navigation inside buildings and allow to gather statistical information about the movement of clients inside shopping malls, airports, etc.

The typical WiFi scan taken using a mobile device or WiFi adapter in mobile robot contains the names of observed WiFi networks, their MAC addresses and corresponding signal strengths in dBm.  Methods to estimate position can be divided into two groups: WiFi ranging~\cite{wifiranging} and WiFi fingerprinting~\cite{radar}.  In WiFi ranging, the properties of WiFi signal wave are exploited to directly estimate the distance to the access points (APs). These solutions work with clear line-of-sight but are impractical inside buildings due to multiple signal occlusions, wall reflections and overall influence of people.  WiFi fingerprinting methods focus on efficiently comparing achieved WiFi scans to the prerecorded database of scans inside buildings and thus are more robust to local signal disturbances.

The processing pipeline for indoor localization systems (ILS) using WiFi fingerprinting is presented based on systems taking part in annual smartphone-based localization challenge at Indoor Positioning and Indoor Navigation conference~\cite{ipin15comp}. The current state-of-the-art WiFi fingerprinting methods assume hierarchical approach to indoor localization, where the captured scan is at first used to precisely estimate the building where the user is located. The most common solution is to associate each Access Point to the building based where the strongest signal of that network was observed in the recorded database. During localization, each network in WiFi scan is analyzed and building is located based on simple voting for each network. Similar procedures are then proposed to estimate proper floor inside the building. The building and floor are usually correctly predicted in $85\%-95\%$ of cases~\cite{ipin15comp}.

The golden standard of estimating user's position in the recognized floor is to use kNN ($k$ nearest neighbors)~\cite{radar} approach, where $k$ scans that are the most similar to the analyzed WiFi scan are queried from the database and their positions are averaged to achieve the position estimate. A modified approach is to use weighted kNN~\cite{wknn}, where the final location is computed similar to kNN, but each queried scan is weighted by the similarity to the current scan. Unfortunately, a dense and precise radio map of WiFi scans recorded every $1-1.5$ m in the environment is needed to achieve precise localization. The acquisition process is time-consuming and the kNN or wkNN algorithm is usually tuned in grid parameter search suited to work in the chosen building~\cite{uji}. Also, a time-consuming process of data analysis is necessary to filter challenging data to increase system precision.

Those solutions are difficult to tune in a case of a larger building and if a large amount of data is available. Nowadays, the growing amount of available data is caused by the popularity of mobile devices equipped with WiFi adapters. Therefore, machine learning approaches are a promising solution due to less parameter tuning and better scalability in larger environments. In recent works, random forests~\cite{wifirobotrandomforest,ipin16randomforest} were already applied with some success. To the best knowledge of the authors, this article is the first presenting the use of deep neural networks (DNN) for WiFi fingerprinting, while the use of DNN for Ultra Wide Band (UWB) data on a carefully recorded dataset is also a very recent idea~\cite{deepbeliefnetworks}.

Therefore, we propose to use deep neural networks with efficient stacked autoencoders~\cite{stackedautoencoders} to reduce feature dimensionality and to take advantage of a large amount of gathered data. We propose DNN approach for floor and building classification problem and evaluate our system on publicly available data. We believe that DNN approach can reduce the necessary workforce needed to tune WiFi systems and will encourage to use WiFi localization subsystems in mobile robots.

In chapter~\ref{sec:systems} we present a detailed architecture of our system. The system introduction is followed by a presentation of experimental data in chapter~\ref{sec:data} that is used for the evaluation performed in chapter~\ref{sec:exp}. Finally, the conclusions are drawn in chapter~\ref{sec:con}.

\section{Low-effort place recognition with deep learning} \label{sec:systems}

Recent developments in the field of deep neural networks revolutionized image analysis achieving better results than previous state-of-the-art algorithms. The requirement is that a large database of training samples has to be available. Luckily, gathering a large amount of WiFi scans in the environment is feasible as it is possible by means of crowdsourcing. The rise of deep neural network popularity can be also attributed to increased GPU processing capabilities and emergence of high-level DNN libraries that ease implementation of complex ideas.

In our DNN approach, we also employ the hierarchical approach as in state-of-the-art systems~\cite{ipin15comp}. We propose a classification DNN with autoencoders to predict floor and building based on a single scan. 
We will shortly present used architectures of DNNs.

 \begin{figure}[thpb!] \centering  \includegraphics[width=0.8\columnwidth]{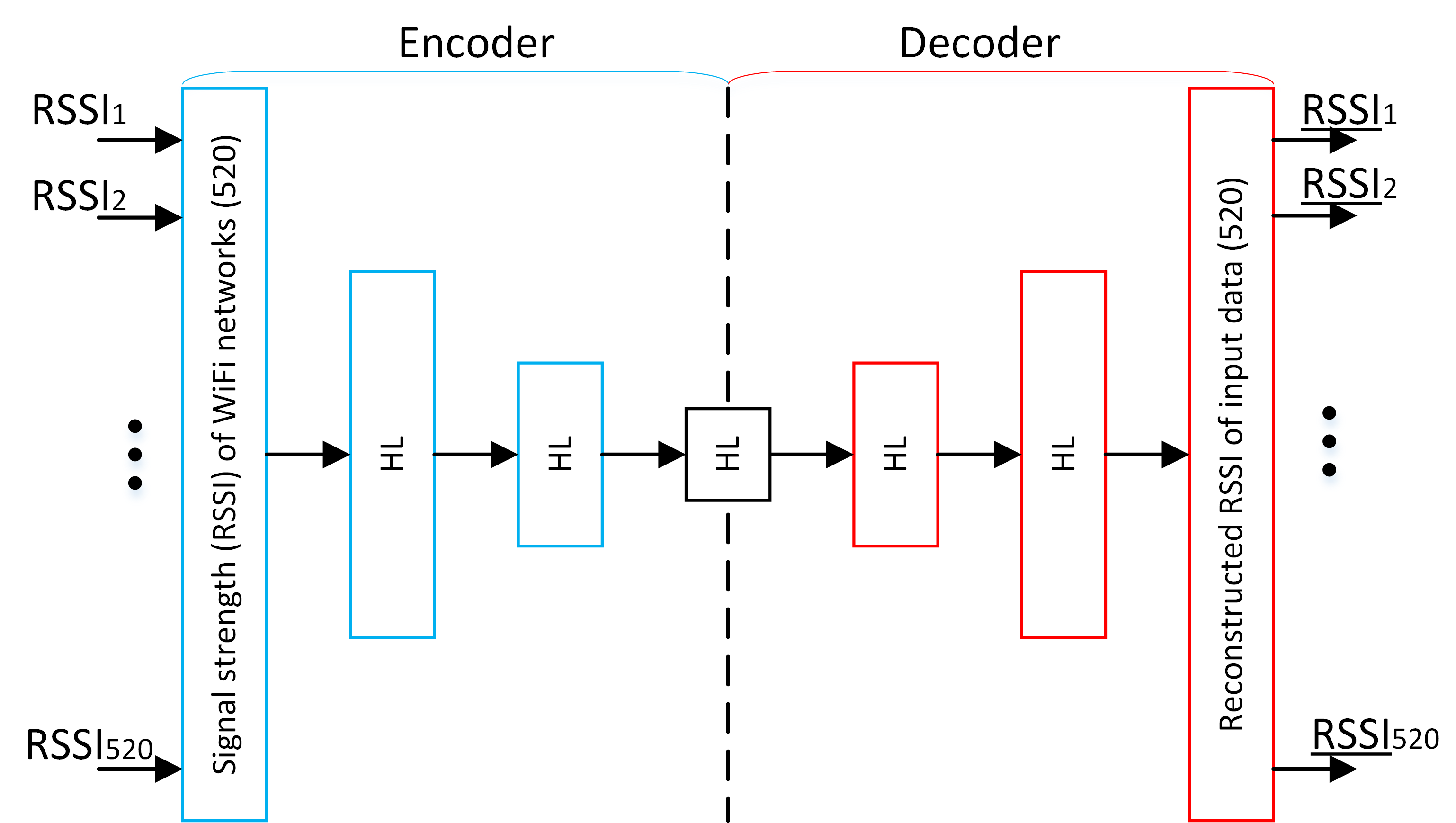}  \caption{Stacked autoencoder (SAE) used in DNN to determine floor and building. The input to SAE are signal strengths detected in a scan with one value for each network visible in the training database. The output of decoder is the reconstructed input from reduced representation. The HL stands for hidden layer and the numbers in parentheses represent the number of neurons in the layer} \label{fig:encoderDecoder} \end{figure}

Each WiFi scan contains the signal strength measurements for APs available in its vicinity, but only a subset of a total number of networks in the environment are observed. Therefore, it is difficult to propose a reduced higher-level features for machine learning approaches. Fortunately, we can use stacked autoencoders~\cite{stackedautoencoders} for this task and provide raw measurements at DNN input. Stacked autoencoders (SAE) are parts of the deep network used to reduce the dimensionality of the input data by learning the reduced representation of the original data during unsupervised training. The used SAE (blue) with additional decoder part (red) is presented in Fig.~\ref{fig:encoderDecoder}. The SAE is learned during unsupervised training and the goal is to train the pair encoder-decoder to achieve the same information at the output as it was provided as input. Due to the fact that the dimensionality of the layer between encoder and decoder is smaller than the size of the input vector, the network has to learn the reduced representation of information provided at the input.

\begin{figure}[thpb!] \centering  \includegraphics[width=0.8\columnwidth]{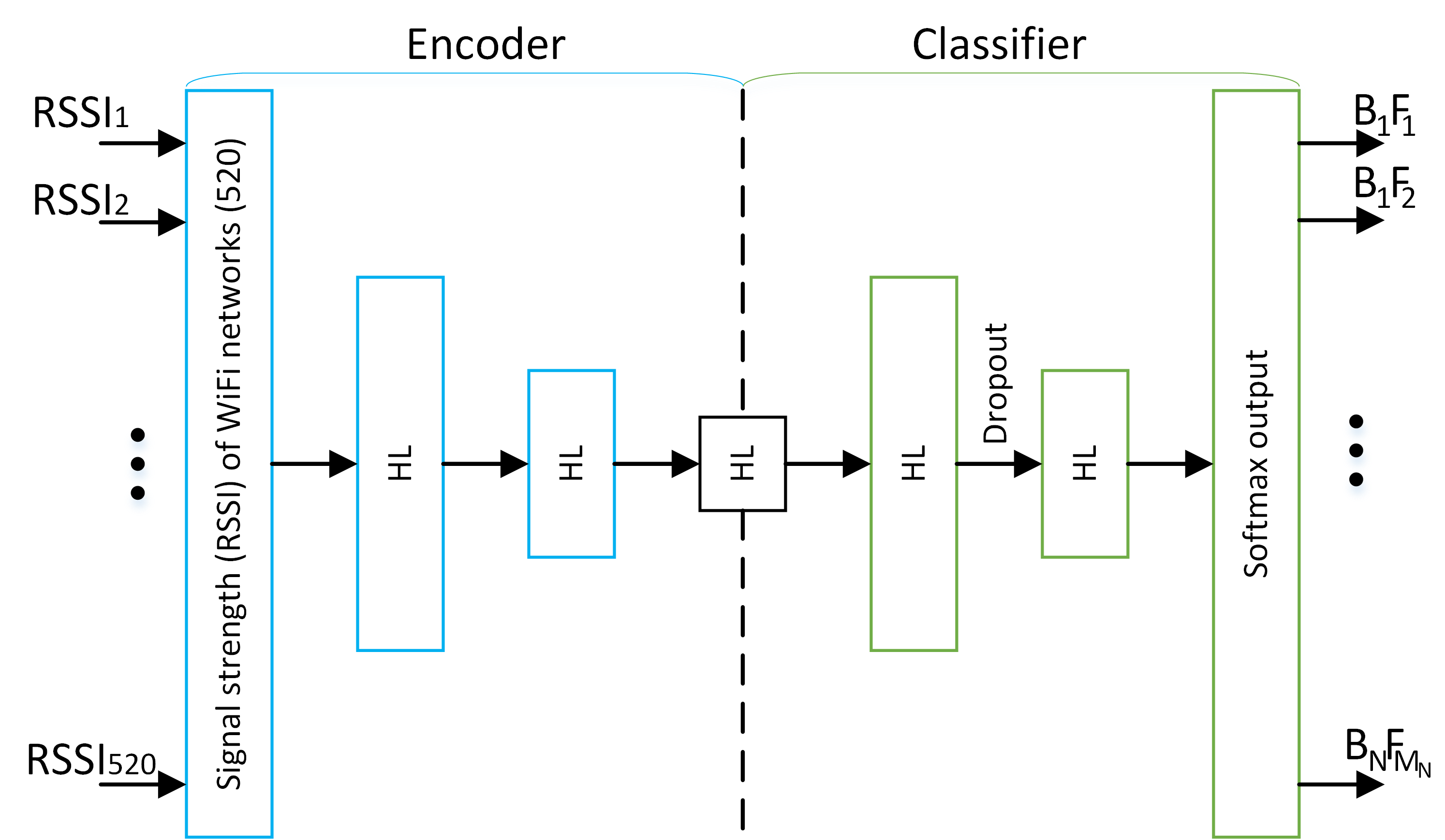}  \caption{Architecture of DNN with SAE used to classify building and floor based on provided input WiFi scan. The already pre-trained encoder part is connected to classifier. The numbers in parentheses represent the number of neurons in the layer} \label{fig:dnnClassifier} \end{figure}

When the unsupervised learning of weights of SAE is finished, the decoder part of the network is disconnected and typical fully-connected layers of a deep network are connected to the output of  the encoder (as presented in Fig.~\ref{fig:dnnClassifier}), that we call the classifier. In Fig.~\ref{fig:dnnClassifier}, the exemplary classifier part consists of two hidden layers, but the number of neurons has to be chosen based on the complexity of the problem.  We also employ dropout~\cite{dropout} between hidden layers of the classifier, which randomly drops connections between layers during training to force the network to learn redundant representation and thus achieve better generalization and avoid overfitting. The final output layer is a softmax layer that outputs the probabilities of a current sample belonging to analyzed classes. If the number of buildings is denoted by $N$ and the number of floors in a building $i$ is denoted by $M_i$, the output layer consists of $\sum_{i=1}^{N} M_i$ neurons.

Before the supervised learning is performed, the labeled data is divided into the training, validation, and testing sets. The DNN is learned on training data with performance checks on validation data, while final performance is provided based on the testing set. During the training process, weights of SAE learned during unsupervised training and weights in classifier are modified to create a final system. In our solution, the training process uses categorical cross entropy error with the Adam optimizer.




In our work we used Keras library\footnote{\url{https://keras.io/}} for deep neural networks with TensorFlow\footnote{\url{https://www.tensorflow.org/}} for numerical computation using data flow graphs and Scikit-learn~\footnote{\url{http://scikit-learn.org/}} for typical machine learning operations.

 \section{Experimental data} \label{sec:data}

To evaluate proposed algorithms and allow comparison with methods proposed by other researchers, a large dataset that contains labeled positions and is publicly available is necessary. Fortunately, there is the UJIIndoorLoc dataset~\cite{uji} that contains WiFi measurements used during EvAAL competition at IPIN 2015~\cite{ipin15comp} and is made publicly available\footnote{\url{https://archive.ics.uci.edu/ml/datasets/UJIIndoorLoc}}. The dataset consists of 21048 WiFi scans that are divided into 19937 training and 1111 validation samples that were recorded in buildings of the University of Jaume I in Spain at the area of almost 110000 square meters. The registered data comes from 25 different Android devices and was recorded with the help of 20 users.

Each scan in the database contains 529 attributes. At the area of operation, 520 different APs were discovered and therefore the first 520 attributes inform about the received signal strength of those networks. The signal strengths vary from -104 dBm in a case of poor reception to almost 0 dBm. When AP is not available, the value of 100 is provided. The remaining 9 attributes contain information about longitude and latitude of measurement, floor number, building ID, space ID, relative position, user ID, phone ID and the timestamp of the measurement.

The public version of  UJIIndoorLoc dataset doesn't contain the testing samples made available only for the competitors at EvAAL~\cite{ipin15comp}. Therefore, we decided to randomly divide UJI training data samples into training and validation samples and treat UJI validation data as testing samples. This operation allows us to reliably compare our results with results obtained by teams participating in mentioned competition.

The UJIIndoorLoc dataset consists of series of user measurements without averaging, no dense radio map is available and validation/testing is performed 4 months later than original data gathering.  Therefore, the localization on data from UJIIndoorLoc is challenging, but obtained results can be used to estimate the real-life performance of the system.

\section{Experimental evaluation} \label{sec:exp}

\subsection{Representing the lack of WiFi signal}

Using machine learning approaches for WiFi fingerprinting is challenging as the number of features (equal to the number of APs) is big comparing to the number of features containing signal strengths of detected networks. Therefore, the problem is to define a representation of the lack of measurement for the majority of features. In the representation used in UJIIndoorLoc, the lack of measurement is denoted by 100, when measured signals strengths vary from $0$ to $-104$ dBm.  In the performed tests, we assumed that the lack of signal strength measurement is equal to $100$ or assumed that it is equal to $-110$ dBm, which is a signal weaker that the weakest signal ever received in the provided dataset. Additionally, we scaled the WiFi measurements to have mean equal to $0$ and variance equal to $1$.  We also tested scaling measurements for each WiFi independently or computed the same scaling for all WiFi measurements. The influence of those parameters for floor and building classification problem with the network consisting of SAE (256-128-64) containing three hidden layers of 256, 128, and 64 neurons, and the classifier (128-128) with two hidden layers containing 128 and 128 neurons was measured and is presented in Tab.~\ref{tab:lackOfMeasurment}.

 \begin{table}[htbp!] \centering \caption{Achieved correct building and floor recognition rates with different input data scaling and treatment of the lack of AP measurement} \label{tab:lackOfMeasurment} \begin{tabular}{c|c|c|c|c} Network used                                                                                             & Lack of AP & Scaling & Validation dataset & Testing dataset \\ \hline  \multirow{4}{*}{\begin{tabular}[c]{@{}c@{}}SAE (256-128-64) \\ + \\ Classifier (128 - 128)\end{tabular}} & 100        & Independent              & 0.973           & 0.529        \\                                                                                                          & -110       & Independent              & 0.996           & 0.851        \\                                                                                                          & 100        & Joint       & 0.995           & 0.845        \\                                                                                                          & -110       & Joint                     & 0.982           & 0.911        \end{tabular} \end{table}

The obtained results on validation dataset are similar, but the best results are obtained when each WiFi measurement is scaled in the same way and lack of AP is represented by a value of -110 dBm. We believe that independent scaling does not provide expected better results due to the fact that some networks are underrepresented and thus the proposed scaling is not suitable for WiFi measurements that are out of the range calculated for that network on the basis of the training samples. The problems with representing the lack of network as 100 dBm might stem from data discontinuity as greater measurements do not mean better reception, which is not the case for representing a lack of measurement as -110 dBm.

\subsection{Building and floor classification}

\begin{figure}[thpb!] \centering  \includegraphics[width=\columnwidth]{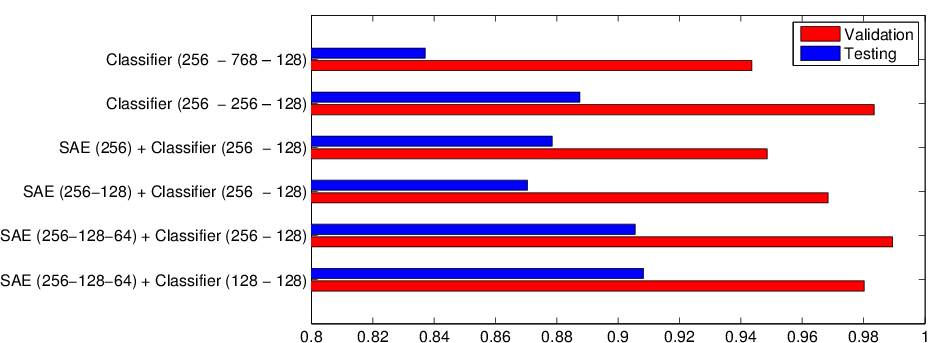}  \caption{Comparison of correct recognition ratios for different DNN architectures evaluated in building and floor classification problem obtained on validation and testing samples.           The numbers in parentheses inform about the number of neurons in the hidden layers.} \label{fig:dnnResult} \end{figure}

When it comes to building and floor classification problem, several DNN architectures with most distinguishable results are presented in Fig.~\ref{fig:dnnResult}. For each architecture, we tried different options, adding dropout layers with a percent of dropped connections varying from $5\%$ to $20\%$ and different values of learning rates for the Adam optimizer to achieve the best results.

The first two architectures in Fig.~\ref{fig:dnnResult} are fully-connected neural networks without SAE part. Those networks still can be considered deep neural networks as these networks utilize dropout to prevent overfitting. Increasing the number of neurons in those networks resulted in increased correct recognitions on validation dataset, even up to $98\%$ correct building and floor detections, but performed significantly worse in a case of independent testing dataset yielding up to $88\%$. We believe that $10\%$ difference in performance on validation and testing dataset is due to the problem that only part of the input vector contains signal strengths of APs and the rest is equal to no signal strength measurement.

Therefore, we decided to utilize SAE to efficiently reduce the dimensionality of input vector from 520 networks to 256, 128 and 64 features in different scenarios. The SAE part was then connected to typical classifier part. The proposed architecture still yields of around $99\%$ of correct recognitions on validation dataset but allowed to increase the correct recognition rate on testing samples up to $91\%$. This proves that SAE learns a simplified representation of input information that enables the system to achieve better results than networks without autoencoders.

Finally, the architecture with the best ratio of correct recognitions, SAE (256-128-64) + Classifier (128-128), was trained with the same parameters on training and validation data.  The recognition test was then performed on the testing samples yielding $92\%$ correct building and floor recognitions. The naive approach that finds the most similar scan in training samples based on Euclidean distance resulted in only $75\%$ correct recognitions. Comparing to results obtain during EvAAL competition, the competing teams achieved results from around $87\%$ to $96\%$ percent, but on the original testing dataset that was unavailable to the authors of the article. Nevertheless, achieved results suggest that the DNN approach might achieve comparable results to state-of-the-art systems that took months to carefully develop and tune.



\section{Conclusions} \label{sec:con}

Applying deep learning to WiFi fingerprinting allowed to achieve a system that estimates floor and building on the publicly available dataset with the accuracy that is comparable to state-of-the-art approaches, but allows to reduce the needed effort as no additional tuning or filtering is needed. The application of stacked autoencoders (SAE) for WiFi feature space reduction results in deep networks that can be easier to learn and perform better on the testing dataset. The future work will focus on using deep learning for estimating the final X, Y position of the user by solving the regression problem, and on applying the proposed DNN architecture for estimating the user position inside smaller clusters/areas defined at each floor.

 \section*{Acknowledgements}

This research was funded by the National Science Centre in Poland in years 2016-2019 under the grant 2015/17/N/ST6/01228.

\bibliographystyle{unsrt} 
\bibliography{Automation}

\begin{thebibliography}{10}

\bibitem{eaai01}
A.~Kasi\'nski and P.~Skrzypczy{\'n}ski.
\newblock Perception network for the team of indoor mobile robots. concept.
  architecture. implementation.
\newblock {\em Engineering Applications of Artificial Intelligence},
  14(2):125--137, 2001.

\bibitem{psamcs}
P.~Skrzypczy{\'n}ski.
\newblock Simultaneous localization and mapping: A feature-based probabilistic
  approach.
\newblock {\em Int. Journal of Applied Mathematics and Computer Science},
  19(4):575--588, 2009.

\bibitem{orbslam}
J.~M. M.~Montiel R.~Mur-Artal and J.~D. Tard\'{o}s.
\newblock {ORB-SLAM: a Versatile and Accurate Monocular SLAM System}.
\newblock {\em CoRR}, abs/1502.00956, 2015.

\bibitem{ecmr13}
M.~Nowicki and P.~Skrzypczy{\'n}ski.
\newblock Combining photometric and depth data for lightweight and robust
  visual odometry.
\newblock In {\em European Conf on Mobile Robots (ECMR)}, pages 125--130,
  Barcelona, 2013.

\bibitem{iciar}
M.~Fularz, M.~Nowicki, and P.~Skrzypczy\'{n}ski.
\newblock Adopting feature-based visual odometry for resource-constrained
  mobile devices.
\newblock In A.~Campilho and M.~Kamel, editors, {\em Image Analysis and
  Recognition}, volume LNCS 7324, pages 431--441. Springer, 2014.

\bibitem{mobicase}
M.~Nowicki and P.~Skrzypczy{\'n}ski.
\newblock Indoor navigation with a smartphone fusing inertial and wifi data via
  factor graph optimization.
\newblock In {\em MobiCASE 2015}, volume LNICST 162, pages 1--19. Springer,
  2015.

\bibitem{wifirobot1}
J.~M. C.~Plaza V.~M.~Olivera and O.~S. Serrano.
\newblock {WiFi Localization Methods for Autonomous Robots}.
\newblock {\em Robotica}, 24(4):455--461, July 2006.

\bibitem{wifirobot2010}
J.~Biswas and M.~{Veloso}.
\newblock {WiFi} localization and navigation for autonomous indoor mobile
  robots.
\newblock In {\em Robotics and Automation (ICRA), 2010 IEEE International
  Conference on}, pages 4379--4384, May 2010.

\bibitem{wifirobot2015}
M.~Sweatt, A.~Ayoade, Q.~Han, J.~Steele, K.~Al-Wahedi, and H.~Karki.
\newblock {WiFi} based communication and localization of an autonomous mobile
  robot for refinery inspection.
\newblock In {\em 2015 IEEE International Conference on Robotics and Automation
  (ICRA)}, pages 4490--4495, May 2015.

\bibitem{wifiranging}
M.~Ciurana, F.~Barcelo-Arroyo, and F.~Izquierdo.
\newblock {A Ranging Method with IEEE 802.11 Data Frames for Indoor
  Localization}.
\newblock In {\em Proceedings of the 2007 IEEE Wireless Communications and
  Networking Conference}, pages 2092--2096, Washington, DC, USA, 2007. IEEE
  Computer Society.

\bibitem{radar}
P.~Bahl and V.~N. Padmanabhan.
\newblock {RADAR: an in-building RF-based user location and tracking system}.
\newblock In {\em INFOCOM 2000. Nineteenth Annual Joint Conference of the IEEE
  Computer and Communications Societies. Proceedings. IEEE}, volume~2, pages
  775--784 vol.2, 2000.

\bibitem{ipin15comp}
A.~Moreira, M.~J. Nicolau, F.~Meneses, and A.~Costa.
\newblock {Wi-Fi fingerprinting in the real world - RTLS@UM at the EvAAL
  competition}.
\newblock In {\em Indoor Positioning and Indoor Navigation (IPIN), 2015
  International Conference on}, pages 1--10, Oct 2015.

\bibitem{wknn}
S.~Khodayari, M.~Maleki, and E.~Hamedi.
\newblock {A RSS-based fingerprinting method for positioning based on
  historical data}.
\newblock In {\em Performance Evaluation of Computer and Telecommunication
  Systems (SPECTS), 2010 International Symposium on}, pages 306--310, July
  2010.

\bibitem{uji}
J.~Torres-Sospedra, R.~Montoliu, A.~Mart\'{i}nez-Us\'{o}, J.~P. Avariento,
  T.~J. Arnau, M.~Benedito-Bordonau, and J.~Huerta.
\newblock {UJIIndoorLoc}: A new multi-building and multi-floor database for
  wlan fingerprint-based indoor localization problems.
\newblock In {\em Indoor Positioning and Indoor Navigation (IPIN), 2014
  International Conference on}, pages 261--270, Oct 2014.

\bibitem{wifirobotrandomforest}
R.~Elbasiony and W.~Gomaa.
\newblock Wifi localization for mobile robots based on random forests and
  gplvm.
\newblock In {\em Machine Learning and Applications (ICMLA), 2014 13th
  International Conference on}, pages 225--230, Dec 2014.

\bibitem{ipin16randomforest}
Y.~Beer.
\newblock {WiFi Fingerprinting using Bayesian and Hierarchical Supervised
  Machine Learning assisted by GPS}.
\newblock In {\em Indoor Positioning and Indoor Navigation (IPIN), 2016
  International Conference on}, Oct 2016.

\bibitem{deepbeliefnetworks}
J.~Luo and H.~Gao.
\newblock {Deep Belief Networks for Fingerprinting Indoor Localization Using
  Ultrawideband Technology}.
\newblock {\em Int. J. Distrib. Sen. Netw.}, 2016:18:18--18:18, January 2016.

\bibitem{stackedautoencoders}
I.~Lajoie Y.~Bengio P.~Vincent, H.~Larochelle and P.-A. Manzagol.
\newblock {Stacked Denoising Autoencoders: Learning Useful Representations in a
  Deep Network with a Local Denoising Criterion}.
\newblock {\em J. Mach. Learn. Res.}, 11:3371--3408, December 2010.

\bibitem{dropout}
N.~Srivastava, G.~Hinton, A.~Krizhevsky, I.~Sutskever, and R.~Salakhutdinov.
\newblock {Dropout: A Simple Way to Prevent Neural Networks from Overfitting}.
\newblock {\em Journal of Machine Learning Research}, 15:1929--1958, 2014.

\end{thebibliography}


\end{document}